\documentclass[letterpaper]{article} 
\usepackage{aaai25}  
\usepackage{times}  
\usepackage{helvet}  
\usepackage{courier}  
\usepackage[hyphens]{url}  
\usepackage{graphicx} 
\usepackage{natbib}  
\usepackage{caption} 
\usepackage{algorithm}
\usepackage{algorithmic}
\usepackage{multirow}
\usepackage{threeparttable}
\usepackage{booktabs}
\usepackage[tight]{subfigure}
\usepackage{xcolor}
\usepackage{amsmath}
\usepackage{array}
\usepackage{makecell}

\usepackage{newfloat}
\usepackage{listings}
\DeclareCaptionStyle{ruled}{labelfont=normalfont,labelsep=colon,strut=off} 
\floatstyle{ruled}
\newfloat{listing}{tb}{lst}{}
\floatname{listing}{Listing}
\title{Evaluating Hallucination in Large Vision-Language Models based on Context-Aware Object Similarities}
\author {
    Shounak Datta$^*$\textsuperscript{\rm 1},
    Dhanasekar Sundararaman$^*$\textsuperscript{\rm 2}
}
\affiliations {
    \textsuperscript{\rm 1}Indian Statistical Institute\\
    \textsuperscript{\rm 2}Duke University\\
    shounak.jaduniv@gmail.com, ds448@duke.edu
}

\begin{document}

\maketitle
\def\thefootnote{*}\footnotetext{These authors contributed equally to this work}\def\thefootnote{\arabic{footnote}}

\begin{abstract}
Despite their impressive performance on multi-modal tasks, large vision-language models (LVLMs) tend to suffer from hallucinations. An important type is object hallucination, where LVLMs generate objects that are inconsistent with the images shown to the model. Existing works typically attempt to quantify object hallucinations by detecting and measuring the fraction of hallucinated objects in generated captions. Additionally, more recent work also measures object hallucinations by directly querying the LVLM with binary questions about the presence of likely hallucinated objects based on object statistics like top-$k$ frequent objects and top-$k$ co-occurring objects. In this paper, we present Context-Aware Object Similarities (CAOS), a novel approach for evaluating object hallucination in LVLMs using object statistics as well as the generated captions. CAOS uniquely integrates object statistics with semantic relationships between objects in captions and ground-truth data. Moreover, existing approaches usually only detect and measure hallucinations belonging to a predetermined set of in-domain objects (typically the set of all ground-truth objects for the training dataset) and ignore generated objects that are not part of this set, leading to under-evaluation. To address this, we further employ language model--based object recognition to detect potentially out-of-domain hallucinated objects and use an ensemble of LVLMs for verifying the presence of such objects in the query image. CAOS also examines the sequential dynamics of object generation, shedding light on how the order of object appearance influences hallucinations, and employs word embedding models to analyze the semantic reasons behind hallucinations. By providing a systematic framework to identify and interpret object hallucinations, CAOS aims to offer a nuanced understanding of both the hallucination tendencies of LVLMs and the factors contributing to object hallucinations.
\end{abstract}

%

\section{Introduction}

Large language models (LLMs) like LLaMA, Gemini, and Mixtral \citep{touvron2023llama, touvron2023llama2, team2023gemini, jiang2024mixtral} have demonstrated remarkable capabilities in natural language processing tasks. Building upon this success, researchers have focused on integrating powerful LLMs with visual encoders to create large vision-language models (LVLMs) \citep{liu2024visual,gong2023multimodal,zhu2023minigpt,dai2024instructblip, ye2023mplug}. These LVLMs leverage the language understanding abilities of LLMs while enhancing their capabilities to process and interpret visual information seamlessly. LVLMs typically utilize the visual encoders to analyze image data, while replacing the original language encoders with state-of-the-art LLMs. Through a combination of vision-language pretraining and fine-tuning on visual instructions \citep{wang2021simvlm}, LVLMs have demonstrated impressive performance on complex tasks that require integrating visual and linguistic information. 

LVLMs exhibit robust proficiency across various vision-language tasks, showcasing their versatility and adaptability. For instance, they excel in tasks such as image captioning \citep{herdade2019image}, generating descriptive textual representations of visual content to bridge the semantic gap between images and language. Additionally, LVLMs demonstrate proficiency in visual question answering \citep{antol2015vqa, wu2017visual}, comprehending and responding to queries posed in natural language based on visual input.

The development of LVLMs marks a significant milestone in the convergence of vision and language modalities, opening up new avenues for research and application in multi-modal models. However, LLMs, and similarly LVLMs, often suffer from the issue of hallucination \citep{rawte2023survey, liu2024survey, xu2024hallucination}, where the generated content contains information that is inconsistent or unfaithful to the provided input data. Hallucination refers to the phenomenon where the generated content contains nonsensical, contradictory or factually incorrect information that violates the input instructions or prompts. A common form of hallucination in LVLMs is object hallucination (Rohrbach et al. 2018; Li et al. 2023; Zhou et al. 2023), where the model describes objects or entities that are not present in the visual input. Recent studies have shown that LVLMs suffer from severe object hallucination issues (Li et al. 2023; Zhou et al. 2023), often generating descriptions that include objects inconsistent with the inputs. Hallucinations can be influenced by the visual instructions or prompts, as objects frequently occurring in the instructions or co-occurring with image objects are more prone to being hallucinated.

\begin{figure*}[!t]
    \hspace*{0.5cm}
    \centering
    \includegraphics[width=0.75\textwidth]{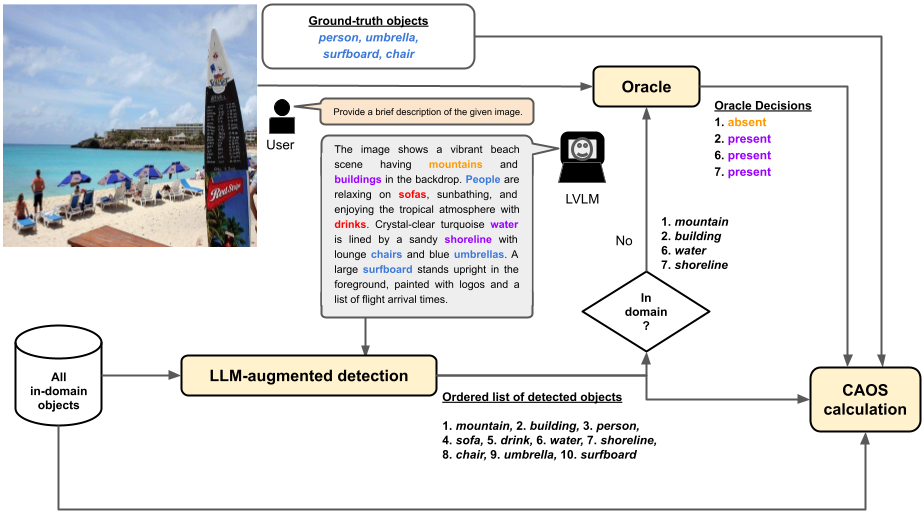}
    \caption{Overview of the CAOS framework: The CAOS framework evaluates LVLMs by generating captions for images with known ground-truth object annotations. The captions may include both in-domain and out-of-domain objects, which could be real or hallucinated. Specific color coding identifies object types: blue for real in-domain objects, red for hallucinated in-domain objects, purple for real out-of-domain objects, and orange for hallucinated out-of-domain objects. The key components of the framework are highlighted in yellow: constructing an ordered list of objects in the caption using an LLM-augmented identification module, querying an oracle (an ensemble of LVLMs) to confirm the presence or absence of out-of-domain objects (isolated on the basis of known in-domain objects in the training dataset) as ground-truth annotations are unavailable, and finally calculating CAOS scores (best viewed in color).} 
    \label{fig:flow}
\end{figure*}


Hallucination in both LLMs and LVLMs can be problematic, as it can lead to the generation of unreliable or misleading information. This issue undermines the reliability of LVLM outputs despite their semantic coherence. Moreover, it also poses potential risks in real-world applications where accurate interpretation of visual content is of paramount importance. Therefore, it has prompted researchers to develop evaluation benchmarks and mitigation techniques to detect and mitigate hallucination in these models. Existing works like \citep{li2023evaluating, pham2024hpope} use object statistics such as top-$k$ objects and frequently co-occurring objects to determine the causes of hallucination and belong to a family of methods which ensure LVLMs respond to a set of predefined questions to evaluate hallucinations \citep{lovenia2023negative}. Others works may focus on a particular aspects of object hallucinations, like numbers \citep{zhang2024evaluating}. Moreover, the object statistics mentioned above can also then be used to mitigate hallucinations by post-hoc rectification \citep{zhou2023analyzing}. In addition to the object statistics, the generated captions also contain additional information about an LVLM's tendency to hallucinate, in the form of semantic relationships between the objects in the generated text. However, existing methods do not attempt to investigate the effect that the interplay between object statistics and the semantics of the objects present either in the query image or the already generated context can have on object hallucinations during the remainder of the generation process. Consequently, in this paper, we present a novel approach named context-aware object similarities (CAOS) for systematically evaluating object hallucination in LVLMs using the generated captions. By focusing on the interplay between generated objects (hallucinated or otherwise), their position in the generated captions, ground-truth objects as well as the object statistics of the training data, CAOS attempts to offers a more nuanced understanding of object hallucination dynamics and encompasses the the following key improvements over existing works:
\begin{itemize}
    \item \textbf{Detect hallucinations of out-of-domain objects:} Unlike existing methods which rely either on rule-based parsing of known in-domain objects \cite{rohrbach2018chair} or on known occurrence statistics of these objects from the training data \cite{li2023evaluating}, we propose a novel way to augment object detection from generated captions using LLMs and to verify the existence of candidate out-of-domain objects in the images using an oracle consisting of an ensemble of LVLMs.

    \item \textbf{Sequential generation dynamics:} Recognizing the sequential nature of caption generation, CAOS investigates how the order of object appearance influences hallucination. By delving into this, our approach sheds light on how the dynamics of object generation impacts object hallucinations in addition to other critical factors, namely ground-truth objects and frequently occurring objects in the training dataset.

    \item \textbf{Semantic reasons behind hallucinations:} Unlike traditional evaluation metrics, CAOS employs word embedding models to scrutinize the semantic relation between hallucinated objects and ground-truth objects, other objects in the generated captions and frequent objects from the training dataset.
\end{itemize}

With an aim to foster the development of more robust and reliable LVLMs for diverse real-world applications, CAOS intend to enrich the existing literature on evaluating object hallucinations by offering a framework for better identifying object hallucinations in LVLMs, quantifying the semantic reasons behind such hallucinations, and for interpreting these results (potentially in tandem with existing metrics such as CHAIR \citep{rohrbach2018chair} and POPE \cite{li2023evaluating}) to obtain a meaningful ranking not just based on an LVLM's affinity to hallucinate but also the factors influencing such hallucinations.

\section{Related Works}
There are a number of related works on evaluating object hallucinations in LVLMs. \citeauthor{rohrbach2018chair} (\citeyear{rohrbach2018chair}) proposed the CHAIR family of metrics which uses rule-based parsing to identify in-domain objects in an LVLM-generated annotation and then calculates the fraction of hallucinated objects per caption and the fraction of annotations with hallucinated objects in a set of generated annotations. Another relevant precursor to our work is POPE \citep{li2023evaluating}. POPE measures the tendency of an LVLM to hallucinate objects by asking yes/no questions to the model about whether certain objects exist in the image, instead of directly evaluating the generated annotations. The choice of objects to be used for such queries can be random, or based on object statistics of the pre-training datasets, such as objects known to co-occur with ground-truth objects, or most frequent objects in the dataset. H-POPE \cite{pham2024hpope} is an extension of POPE that assesses hallucinations in object existence as well as attributes by hierarchically refining the yes/no questions from coarse-to-fine-grained, progressively probing about the attributes of the objects in the image. \citeauthor{lovenia2023negative} (\citeyear{lovenia2023negative}) proposed NOPE, that uses a static set of negative pronouns to determine if a model hallucinates. There are a couple of works that detect object hallucination and perform post-hoc tuning to generate captions that do not contain hallucinations \cite{zhou2023analyzing, dai2022plausible}. 

\begin{figure*}[!t]
    \centering
    \hspace*{5mm}
    \includegraphics[width=0.75\textwidth]{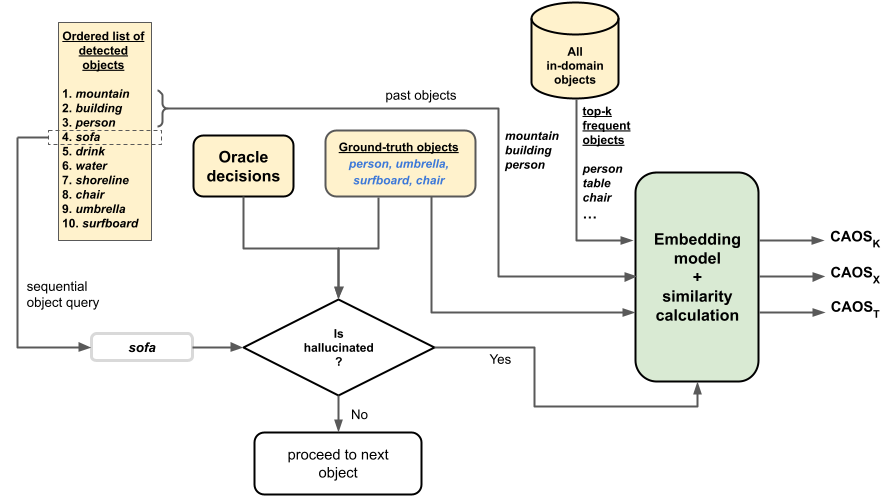}
    \caption{CAOS scores are calculated for in-domain and out-of-domain hallucinated objects which are respectively identified from the ordered list of generated objects using ground-truth annotations and oracle decisions (about presence or absence in the image). CAOS$_T$, CAOS$_X$, and CAOS$_{K}$ are calculated as the maximum cosine similarity between the embeddings of the hallucinated object and those of ground-truth objects, preceding objects in the generated caption, and top-$k$ frequent objects in the training dataset, respectively. Inputs to the CAOS calculation module are highlighted in yellow for clarity (best viewed in color).} 
    \label{fig:caos_calculation}
\end{figure*}

\section{Background}

\subsection{Large Vision-Language Models}
An LVLM architecture comprises of a vision encoder, a language encoder (i.e., an LLM), and a cross-modal alignment network. The vision backbone and cross-modal alignment networks are usually used to convert an input image $I$ into a set of $n$ visual tokens $X_{1:n}$, whereas the accompanying text query is converted to text tokens $Y_{1:q}$ and the subsequent response by the LVLM, parameterized by weights $\theta$, is generated based on the probabilities
\begin{equation}
    p(Y_{q+1:q+r}) = \prod_{t=q+1}^{q+r} p_{\theta}(Y_t|Y_{<t}, X_{1:n}).
\end{equation}

The training pipeline of LVLMs entails several crucial steps:

\textbf{Pre-training on Unimodal Data:} Initially, the vision encoder and the language encoder undergo pre-training on large-scale unimodal datasets, encompassing image and text data, respectively.

\textbf{Image-Text Alignment Pre-training:} Subsequently, these encoders are aligned through image-text alignment pre-training. This alignment enables LVLMs to generate coherent and meaningful captions for given images by leveraging the synergy between visual and textual modalities.

\textbf{Fine-tuning on Image-Text Instructions:} The aligned LVLM model undergoes further fine-tuning on image-text instructions to refine its ability to generate satisfactory answers to natural language questions related to specific images. This fine-tuning process enhances the model's performance on diverse multimodal tasks.

\subsection{Object Hallucination}
Object hallucination occurs when a LVLM references objects that are not actually present in the image it is describing. This can produce unexpected outcomes, particularly in applications like visual question answering, image captioning, or event detection, where accurate identification of objects is crucial. Several factors can contribute to these hallucinations:

\textbf{Common Objects in Training Data:} LVLMs might hallucinate objects that are frequently represented in their training datasets. Examples of such objects include ``person," ``dining table," and ``chair," which are prevalent in visual instruction datasets. 

\textbf{Co-occurring Objects:} Hallucinations may also arise when LVLMs predict objects that commonly appear together with the actual objects in the image, based on patterns observed in the training data.

\textbf{Impact of Visual Instructions:} The type of visual instructions used during training plays a significant role in hallucination tendencies. For instance, LVLMs like InstructBLIP, which are trained on diverse public datasets with concise instructions, are more likely to generate accurate, albeit brief responses. In contrast, models trained with extended synthetic instructions from unimodal language models may be prone to hallucinating due to inconsistencies or excessive detail in the synthetic data.

\textbf{Influence from the already generated context:} In addition to the above factors which are known to impact object hallucinations, we also postulate that the objects already mentioned in the prior context at any given point during generation may also cause hallucinations in a manner similar to the co-occurrence statistics of the ground-truth objects mentioned above. On probing LLaVA \citep{liu2024visual} and mPLUG-Owl \cite{ye2023mplug} with the instruction \textit{``Provide a brief description of the given image."} for a subset of 2000 images from the MSCOCO \citep{cocodataset} validation set (identical to that used by POPE \cite{li2023evaluating}), we find that respectively 20\% and 16\% of the hallucinated objects happen to be the most frequent object to have co-occurred in the training dataset with at least one preceding objects in the already generated part of the annotation. Thus, the past objects in the already generated part of an annotation can also influence object hallucinations in the remainder of the annotation which is yet to be generated. While it is straightforward to measure the fraction of hallucinated objects that fall in this category for known in-domain objects, we can not do the same for out-of-domain objects (i.e. objects not belonging to any of the labeled classes in the training data), due to the lack of co-occurrence information.

\section{CAOS: Context-Aware Object Similarities}

\begin{algorithm}[!ht]
    \footnotesize
    \caption{Calculating CAOS}\label{alg:caos}
    {\bfseries Inputs:} LVLM $p_{\theta}$ to be evaluated. \\
    Image $I$ with ground-truth object labels. \\
    Set $G$ of known objects from fine-tuning dataset labels \\  
    Set $K$ of top-$k$ frequent objects from fine-tuning dataset \\
    Rule-based object parser $P$ \\ 
    LLM $q_{\phi}$ for object identification and oracle $O$ \\
    Embedding model $E$ \\
    Cosine similarity operator $S_{cos}$ \\
    \textbf{Outputs:} Scores CAOS$_T$, CAOS$_X$, CAOS$_{K}$, CAOS$_{T/X}$, CAOS$_{X/K}$, and CAOS$_{avg}$. \\ \vspace{-2mm}
    \let \oldnoalign \noalign
    \let \noalign \relax
    \midrule
    \let \noalign \oldnoalign

    \begin{algorithmic}[1]
    \STATE Generate caption for $I$ using $p_{\theta}$.
    \STATE Identify list $L_{1}$ of objects in the generated caption belonging to $G$ using $P$.  
    \STATE Identify list $L_{2}$ of objects in the generated caption not in $G$, using $q_{\phi}$.  
    \STATE Combine lists $L = L_{1} \cup L_{2}$ preserving the order of the objects in the generated caption.
    \STATE Construct map $M: L \rightarrow \{0,1\}$ labeling genuine objects as $1$ and hallucinated objects as $0$ using ground-truth labels for $l \in L_{1}$ and oracle $O$ for $l \in L_{2}$.
    \STATE Initialize lists $T$ and $X$ with ground-truth objects.
    \STATE Expand sets $T = T \cup T'$ and $X = X \cup T'$, where $T' = \{l \in L_2 | M(l)=1\}$. 
    \STATE Initialize empty set $H$, $T_S$, $X_S$, and $K_S$.
    \FOR{all $l \in L$ in order}
        \IF{$M(l)=0$}
            \STATE Add $l$ to set $H$.
            \STATE Add $\max_{o \in T} S_{cos}(E(l),E(o))$ to set $T_S$. \\
            \STATE Add $\max_{o \in X} S_{cos}(E(l),E(o))$ to set $X_S$. \\
            \STATE Add $\max_{o \in K} S_{cos}(E(l),E(o))$ to set $K_S$. \\
        \ENDIF
        \STATE Add $l$ to set $X$.\\
    \ENDFOR
    \STATE Calculate CAOS$_T$ $= \sum T_S/|H|$. 
    \STATE Calculate CAOS$_X$ $= \sum X_S/|H|$. 
    \STATE Calculate CAOS$_{K}$ $= \sum K_S/|H|$.
    \STATE Calculate CAOS$_{T/X}$= CAOS$_T$/CAOS$_X$.
    \STATE Calculate CAOS$_{X/K}$= CAOS$_X$/CAOS$_{K}$.
    \STATE Calculate CAOS$_{avg}$= {\footnotesize $(\text{CAOS}_T + \text{CAOS}_X + \text{CAOS}_{K})/3$}.
\end{algorithmic}
\end{algorithm}

To account for all the different factors that can influence object hallucinations, we propose a set of evaluation metrics based on context-aware object similarities, called CAOS, to holistically measure how the contents of the image, the sequential ordering of generated objects, as well as the dominant contents of the training dataset influences hallucination. The proposed group of measures consists of CAOS$_T$, CAOS$_X$, CAOS$_{K}$, CAOS$_{T/X}$, CAOS$_{X/K}$, and CAOS$_{avg}$. For a given word embedding model $E$, 
the CAOS$_T$, CAOS$_X$ and CAOS$_K$ scores measure the maximum cosine similarity between the embeddings of a hallucinated object and a set of other objects. In particular, CAOS$_T$ measures the maximum cosine similarity for a given hallucinated object with all the ground-truth objects present in the image. CAOS$_X$, on the other hand, measures the maximum cosine similarity between a hallucinate object and all past objects appearing before itself in the generated caption. 
In practice, we also consider all ground-truth objects, irrespective of their position in the generated caption, to be valid past objects for all hallucinated objects, since the ground-truth objects may always influence generation due to their presence in the image. Finally, since it is well-known that the most frequent objects present in the training datasets can also influence hallucinations \cite{li2023evaluating}, CAOS$_{K}$ measures the maximum cosine similarity between the embeddings of the hallucinated object and the top-$k$ frequent objects in the training dataset (MSCOCO in our experiments). 

Furthermore, since it is often relatively more tolerable for hallucinations to be semantically related to the ground-truth objects known to be present in the image than to other objects in the generated caption, we calculate CAOS$_{T/X}$ to be the ratio between CAOS$_T$ and CAOS$_X$. A high CAOS$_{T/X}$ signifies that the hallucinations are mostly influenced by ground-truth objects. Similarly, it may be relatively more tolerable for the hallucinations to be related to past objects in the generated caption (including all ground-truth objects) than to frequent objects in the training dataset which may not have any relation to the contents of the image being described. Therefore, a high value of CAOS$_{T/K}$, which is the ratio between CAOS$_X$ and CAOS$_{K}$, may be desirable. Lastly, we also calculate CAOS$_{avg}$ which is the mean of CAOS$_T$, CAOS$_X$, and CAOS$_{K}$. We argue that a high value of CAOS$_{avg}$ is also desirable in most cases. This is because high CAOS$_{avg}$ values denote that the hallucinations can be accounted for by the mentioned factors and is less likely to have been caused by unknown eccentricities of the LVLM. 

Given an image, the associated ground-truth object labels, and a caption generated by the LVLM for the image, we begin by identifying all objects in the captions. We then identify which of the objects are hallucinated based on the ground-truth object labels (or using an oracle for out-of-domain objects). In order of their appearance in the generated caption, we calculate the CAOS scores for all the hallucinated objects. Consequently, for a given caption containing multiple hallucinated objects, we report average values for all the CAOS scores over all hallucinated objects in the caption. 

\subsubsection{Augmenting object identification using LLM} 
We observe that the standard rule-based parsing which is used by existing methods like CHAIR \cite{rohrbach2018chair} to identify objects in a caption fails to identify objects beyond the set of known objects in the training dataset. Moreover, we observed that the rule-based approach may sometimes fail to even identify the ground-truth objects present in the caption. This can lead to lower recall scores for the LVLM being evaluated when ground-truth objects are missed. On the other hand, this can also lead to under-evaluation of object hallucinations when hallucinated objects are not detected. Therefore, we augmented the rule-based method with an additional list of objects identified by an LLM (LLaMA-2-7B \cite{touvron2023llama2} in our experiments), based on the generated caption, aided by in-context learning on a handful of examples (5-shot examples in our experiments). We also tally the detected objects in this augmented list with the actual words in the caption to weed-out any potential hallucinations by the LLM.

We also evaluate the efficacy of the LLM-augmented object detection by manually extracting objects from the captions generated by LLaVA on a randomly chosen subset of 100 MSCOCO \cite{cocodataset} validation images and comparing these with the objects detected by the rule-based approach and that of the LLM-augmented approach. Due to its rule-based nature, the former approach has perfect precision (i.e. it never detects objects that are not present in the caption). However, we found that the rule-based approach returns an empty list of objects for two captions. On the flip side, this approach is prone to missing objects present in the generated captions, resulting in a recall of 59\%. The LLM-augmented approach, on the other hand, has perfect recall and almost perfect precision at 97\%. The slight dip in precision occurs due to a handful of instances where the LLM-augmented detection method mistook adjectives to be objects, such as ``night" in ``night scene". However, such rare misdetections are further corrected for as a side-effect of the next stage of our process, where we use an oracle to ascertain whether a detected object actually exists in the given image.

\begin{table*}[!ht]
\centering
\begin{threeparttable}
\small
\begin{tabular}{l|ccccc} \toprule
Model & InstructBLIP & LLaVA & mPLUG-Owl & MiniGPT-4 & Multimodal-GPT \\ \midrule
Precision   & \textbf{0.98} 
                   & 0.85 & 0.79 & 0.92 & 0.88 \\
Recall      & 0.62 & \textbf{0.85} 
                          & 0.74 & 0.78 & 0.67 \\
\# Objects  & 2.22 & 4.97 & 4.49 & \underline{4.91} 
                                        & 3.26 \\ 
CHAIR$_S$   & \textbf{0.04} 
                   & 0.51 & 0.56 & 0.33 & 0.32 \\ 
POPE-F1     & \textbf{0.84} 
                   & 0.68 & 0.67 & 0.74 & 0.67 \\ \midrule

CAOS$_{T}$--GloVe        & 0.30 & 0.38 & 0.37 & 0.40 & 0.40 \\
CAOS$_{X}$--GloVe        & 0.32 & 0.41 & 0.41 & 0.45 & 0.42 \\
CAOS$_{K}$--GloVe ($k$=3)     & 0.52 & 0.50 & 0.51 & 0.47 & 0.47 \\ \midrule
CAOS$_{T/X}$--GloVe      & 0.94 & 0.93 & 0.90 & 0.89 & \textbf{0.95} \\
CAOS$_{X/K}$--GloVe    & 0.62 & 0.82 & 0.80 & \textbf{0.96} 
                                                    & 0.89 \\
CAOS$_{avg}$--GloVe      & 0.38 & 0.43 & 0.43 & \textbf{0.44} 
                                                    & 0.43 \\ \midrule \midrule
                         
CAOS$_{T}$--MiniLM-L6     & 0.26 & 0.40 & 0.37 & 0.35 & 0.42 \\
CAOS$_{X}$--MiniLM-L6     & 0.27 & 0.43 & 0.40 & 0.40 & 0.45 \\
CAOS$_{K}$--MiniLM-L6 ($k$=3)   & 0.52 & 0.54 & 0.53 & 0.45 & 0.51 \\ \midrule
CAOS$_{T/X}$--MiniLM-L6   & \textbf{0.96} 
                               & 0.93 & 0.92 & 0.88 & 0.93 \\
CAOS$_{X/K}$--MiniLM-L6 & 0.52 & 0.80 & 0.75 & \textbf{0.89} 
                                                    & 0.88 \\
CAOS$_{avg}$--MiniLM-L6   & 0.35 & \textbf{0.46} 
                                      & 0.43 & 0.40 & \textbf{0.46} \\ \bottomrule
\end{tabular}
\end{threeparttable}
\caption{Average Precision, Recall, the number of objects per generated caption, CHAIR$_S$, POPE-F1, and CAOS scores using GloVe as well as MiniLM-L6 embeddings across 5 different LVLMs.}
\label{tab:main_inst1_2}
\end{table*}

\subsubsection{Oracle using an ensemble of LVLMs}
Since we have no ground-truth reference to identify whether the additional out-of-domain objects identified using the LLM are actually present in the image, we further label these identified objects as either genuine or hallucinated based on an oracle consisting of an ensemble of LVLMs. Given an object and the original image, we query each of the LVLMs in the ensemble with a query of the form \textit{``Does the image contain $<$object$>$? Please respond with only Present or Absent."}, to force the model to vote on the presence or absence of the object in the image. For our experiments, we use an ensemble of InstructBLIP \cite{dai2024instructblip}, LLaVA-7B \cite{liu2024visual}, mPLUG-Owl \cite{ye2023mplug}, and MiniGPT-4 \cite{zhu2023minigpt}, and break ties in favor of absence to minimize false positives and penalize potential hallucinations. 

\begin{figure}[!b]
    \centering
    \includegraphics[width=0.45\textwidth]{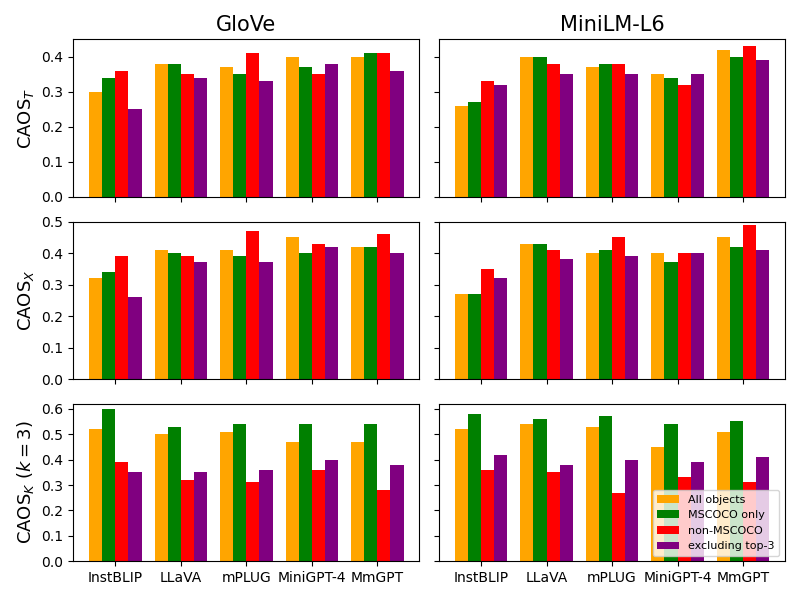}
    \caption{Comparison of CAOS$_T$, CAOS$_X$, and CAOS$_{K}$ scores for all hallucinated objects, only MSCOCO in-domain hallucinated objects, non-MSCOCO hallucinated objects, and all objects barring the top-3 most frequent MSCOCO in-domain objects (best viewed in color).} 
    \label{fig:comparisons}
\end{figure}

We also conduct a manual inspection of a randomly chosen subset of 100 MSCOCO \cite{cocodataset} validation images and the corresponding captions generated by MultimodalGPT \cite{gong2023multimodal} (which is not part of the ensemble) to evaluate the accuracy of the ``Present"/``Absent" decisions made by the ensemble as well as the individual models. We find that the ensemble-based oracle is able to correctly identify the presence or absence of 93.43\% of the 259 non-MSCOCO objects in these images. Among the individual models, InstructBLIP was found to have the highest accuracy at 89.57\%, followed closely by LLaVA at 88.42\%, both MiniGPT-4 and mPLUG-Owl perform slightly worse at 84.94\%. We also illustrate the performance of the oracle with some visual examples in the Appendix. While this may seem similar to querying the model in POPE \cite{li2023evaluating}, it should be noted that the ensemble is likely to misdetect the presence or absence of a smaller number of objects than the individual model being evaluated in POPE. However, these rare misdetections of non-MSCOCO objects may slightly perturb the average CAOS scores because out-of-domain objects tend to have higher CAOS$_T$ and CAOS$_X$ values and lower CAOS$_K$ values compared to known in-domain MSCOCO objects (see Figure \ref{fig:comparisons}).

The complete method is detailed in Algorithm \ref{alg:caos} while Figure \ref{fig:flow} illustrates the overall framework and Figure \ref{fig:caos_calculation} explains the flow of CAOS score calculation.

\begin{figure}[!h]
    \centering
    \includegraphics[width=0.4\textwidth]{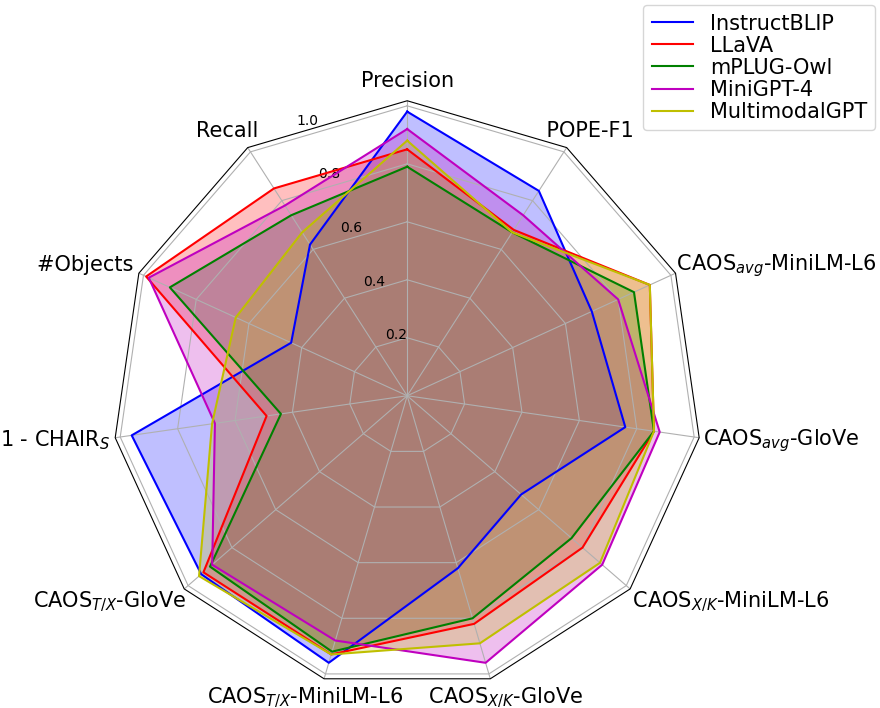}
    \caption{Comparison between all evaluated LVLMs on Precision, Recall, the average number of objects per generated caption (normalized with division by 5), 1 - CHAIR$_S$ (greater is better), POPE-F1, and CAOS$_{T/X}$, CAOS$_{X/K}$, as well as CAOS$_{avg}$ scores (normalized using multiplication by 2) using both GloVe and MiniLM-L6 embeddings (best viewed in color).} 
    \label{fig:radar comparison}
\end{figure}

\section{Results}

We conduct all our experiments on the subset of 2000 images from the MSCOCO \cite{cocodataset} validation set identical to that used by POPE \cite{li2023evaluating}. For each of these 2000 images, we use the instructions \textit{``Provide a brief description of the given image."} and \textit{``Question: Generate a short caption of the image. Answer: "} to probe 5 LVLMs, namely InstructBLIP \cite{dai2024instructblip}, LLaVA \cite{liu2024visual}, mPLUG-Owl \cite{ye2023mplug}, MiniGPT-4 \cite{zhu2023minigpt}, and MultimodalGPT \cite{gong2023multimodal} to generate captions for the image. A detailed comparison of the LVLMs, in terms of model sizes and training recipes is shown in the Appendix.

In Table \ref{tab:main_inst1_2}, we report all 6 CAOS scores, namely CAOS$_T$, CAOS$_X$, CAOS$_{K}$, CAOS$_{T/X}$, CAOS$_{X/K}$, and CAOS$_{avg}$, using two different word embedding models, GloVe \cite{pennington2014glove} and MiniLM-L6 \cite{wang2020minilm, minilm}. We choose these embedding models as they form embeddings based on slightly different objectives. GloVe aims to assign similar embeddings to objects which co-occur is a corpus of documents, while MiniLM-L6 assigns semantically meaningful embeddings based on pretraining on a large number of language tasks such as question answering, natural language inference, etc. For the CAOS$_{K}$ scores, we choose $k=3$ based on the trends of CAOS$_{K}$ scores across $k$ values (see a full discussion in the subsequent section). Additionally, we also report precision (i.e., the fraction of detected objects which are not hallucinations) and recall (which is the fraction of actual ground-truth objects from the query image that are mentioned in the generated caption). Due to the relative instability of the precision measure (related to CHAIR$_I$ \cite{rohrbach2018chair}), we also report POPE-F1 scores for all the models \citep{li2023evaluating}. We also report the fraction of captions having hallucinated objects, which is equivalent to the CHAIR$_S$ metric proposed by \citeauthor{rohrbach2018chair} (\citeyear{rohrbach2018chair}). LVLMs which are prone to generating shorter captions with fewer objects consequently have less scope for hallucination. However, such models also have limited ability to generate faithful descriptive captions for the images. Ideally, a model should have the ability to generate as many objects as needed to generate a particular image while minimizing hallucinations. Therefore, we have also reported the average number of objects generated per caption, as a loose proxy for the capacity of an LVLM to generate articulate captions. 

Overall, the CAOS scores trend similarly across the LVLMs for GloVe and MiniLM-L6. All of the LVLMs exhibit higher CAOS$_{K}$ scores relative to the CAOS$_T$ and CAOS$_X$ scores, implying that all of these models have a high tendency to hallucinate verbatim the frequent objects from the training dataset (i.e. MSCOCO). We also observe a trade-off between the CAOS$_T$ scores (or consequently the related CAOS$_X$ scores) and the CAOS$_{K}$ scores. In other words, LVLMs like InstructBLIP, LLaVA and mPLUG-Owl which have higher CAOS$_{K}$ scores tend to have lower CAOS$_T$ (and CAOS$_X$) scores. This hints at the fact that these models have a relatively higher tendency to hallucinate common objects from the training dataset while the remainder of the models, viz. MiniGPT-4 and MultimodalGPT, have a higher tendency for hallucinations related to the ground-truth objects and the preceding objects in the generated response. 

In order to better compare the performance of the different LVLMs, we also create a radar plot in Figure \ref{fig:radar comparison}, excluding the CAOS$_T$, CAOS$_X$, and CAOS$_{K}$ scores (since those scores do not offer a straightforward way to compare the models, which can be done instead with the other CAOS scores). InstructBLIP exhibits the highest 1-CHAIR$_S$ (or equivalently the lowest CHAIR$_S$) value as well as the highest precision and POPE-F1 scores, implying that it hallucinates less than the other models. However, this is in part due to its tendency to generate a lower number of objects per caption which also results in low recall. Moreover, InstructBLIP also has low CAOS$_{X/K}$ scores. On the other hand, MiniGPT-4 appears to overall have competitive performance across most of the axes, suggesting that it hallucinates less compared to most of the other contenders and that a relatively greater fraction of it's hallucinations are related to the ground-truth objects actually present in the image (a somewhat tolerable property for hallucinations that do happen).

\begin{figure}[!t]
    \centering
    \subfigure[\label{fig:knee1}]{\includegraphics[width=0.23\textwidth]{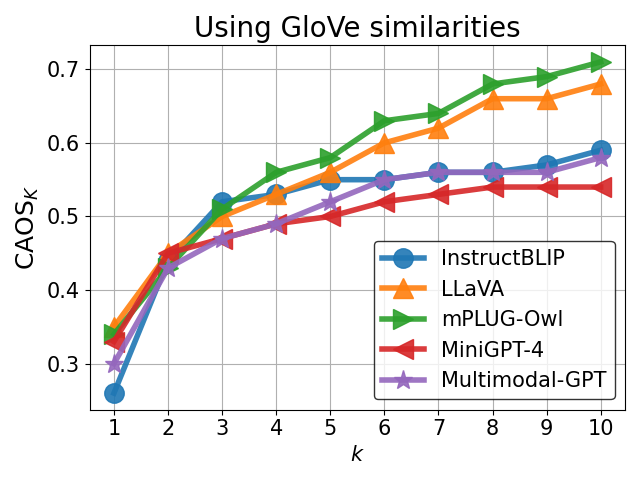}}
    \subfigure[\label{fig:knee2}]{\includegraphics[width=0.23\textwidth]{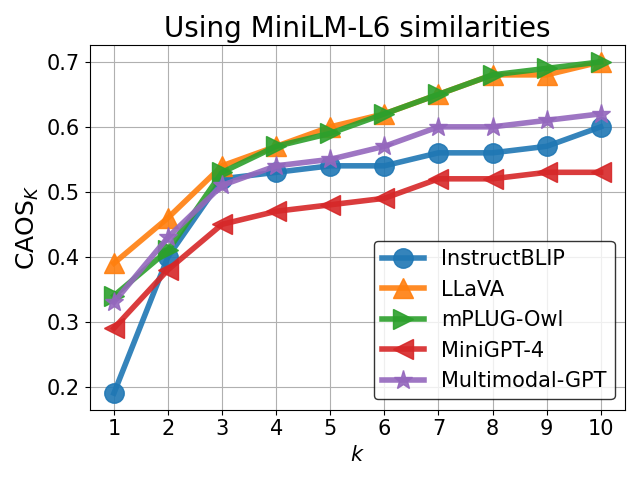}}
    \caption{Trend of CAOS$_{K}$ scores with $k$ varying from 1 to 10 (best viewed in color).}
    \label{fig:topk_knee}
\end{figure}

\subsection{Effect of varying $k$ on CAOS$_{K}$}

The choice of $k$ for CAOS$_{K}$ can affect the CAOS scores. Therefore, we vary $k$ to observe how CAOS$_{K}$ is impacted for the LVLMs. The CAOS$_{K}$ scores for all the LVLMs are shown in Figure \ref{fig:topk_knee}. We observe that the CAOS$_{K}$ scores for all models saturate to some extent at $k=3$, suggesting that the top-3 most frequent objects in MSCOCO disproportionately appear as hallucinations for all the models. This suggests that a choice of $k=3$ can capture most of the information about how the frequent in-domain objects can affect object hallucinations. Additionally, despite the diminishing rate of impact beyond $k=3$, we also observe that LLaVA and mPLUG-Owl continue to be impacted more by frequent objects beyond the top 3.

\subsection{Comparison between different subsets of objects}

In Figure \ref{fig:comparisons}, we investigate how the CAOS$_T$, CAOS$_X$, and CAOS$_{K}$ scores vary across different subsets of hallucinated objects. Since our proposed LLM-augmented object detection is meant to uncover the hallucination of out-of-domain objects, we inspect what the CAOS scores look like for just the hallucinated in-domain MSCOCO objects and just the out-of-domain (denoted as non-MSCOCO) objects. The CAOS scores for both groups largely follow a similar trend to that of all objects, but the in-domain MSCOCO objects seems to have a more pronounced influence from the frequently occurring objects, as implied by the elevated CAOS$_{K}$ scores. Conversely, the out-of-domain objects have a lower albeit non-negligible impact from the frequent MSCOCO objects. This suggests that even the hallucinated out-of-domain objects are semantically influence by the frequent MSCOCO objects to a certain extent. Further, due to the higher impact from the top-3 frequently occurring objects, we also investigate how the CAOS scores change when the hallucinated instances of only the 3 most common MSCOCO objects are excluded. A similar dip in CAOS$_{K}$ is seen for hallucinated objects excluding the top 3, signifying that the top-3 objects disproportionately appear as hallucinations for all models.

\subsection{Consistency across different prompt styles}

\begin{figure}[!t]
    \centering
    \includegraphics[width=0.42\textwidth]{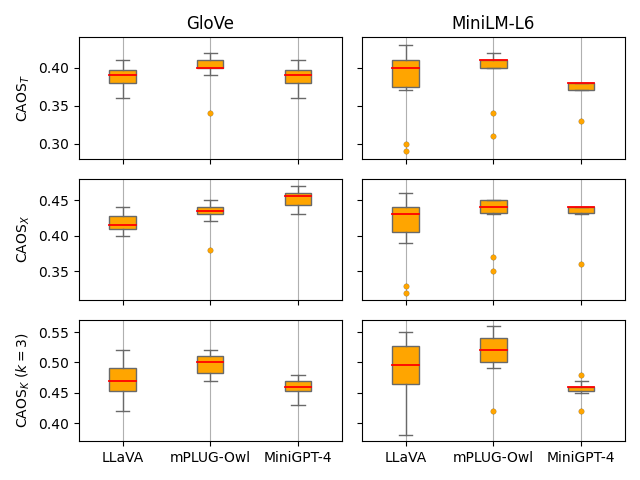}
    \caption{Variability of CAOS scores for LLaVA, mPLUG-Owl, and MiniGPT-4 across 14 different instructions with averages shown in red (best viewed in color).} 
    \label{fig:variability}
\end{figure}

To analyze the sensitivity of CAOS scores to changes in the text prompt, we rerun our experiments on LLaVA, mPLUG-Owl, and MiniGPT-4 with 12 new instructions in addition to the 2 instructions already used in Table \ref{tab:main_inst1_2}. The list of all instructions is shown in the Appendix. We find that the average results over all 14 instructions, detailed in Table \ref{tab:14_instructions}, are overall quite similar to those in Table \ref{tab:main_inst1_2}, indicating that CAOS scores are largely stable to changes in instructions. We also illustrate how CAOS$_T$, CAOS$_X$ and CAOS$_{K}$ values vary across the instructions in Figure \ref{fig:variability}. For a given model and specific CAOS score, we observe some variability across instructions, which is to be expected. Furthermore, while the CAOS scores have different ranges across the different LVLMs, the ranges maintain an ordering similar to that exhibited by the mean CAOS scores (shown in red), further indicating stability across instructions. Finally, the CAOS scores with MiniLM-L6 embeddings look slightly different than those using GloVe. CAOS scores calculated using MiniLM-L6 embeddings seem to be slightly more prone to having outliers than their corresponding GloVe counterparts and CAOS scores for LLaVA have higher variance than the other two models with MiniLM-L6 embeddings.

\begin{table}[!t]
\centering
\begin{threeparttable}
\small
\begin{tabular}{l|>{\centering}p{1cm}>{\centering}p{1cm}c} \toprule
Model           & LLaVA & mPLUG-Owl & MiniGPT-4 \\ \midrule

CAOS$_{T}$--GloVe        & 0.39 & 0.40 & 0.39 \\
CAOS$_{X}$--GloVe        & 0.42 & 0.43 & 0.45 \\
CAOS$_{K}$--GloVe ($k$=3)      & 0.47 & 0.50 & 0.46 \\ \midrule
CAOS$_{T/X}$--GloVe      & \textbf{0.93} 
                               & \textbf{0.93} 
                                      & 0.87 \\
CAOS$_{X/K}$--GloVe    & 0.89 & 0.86 & \textbf{0.98} \\
CAOS$_{avg}$--GloVe      & 0.43 & \textbf{0.44} 
                                      & 0.43 \\ \midrule \midrule

CAOS$_{T}$--MiniLM-L6     & 0.39 & 0.40 & 0.37 \\
CAOS$_{X}$--MiniLM-L6     & 0.41 & 0.43 & 0.43 \\
CAOS$_{K}$--MiniLM-L6 ($k$=3)   & 0.49 & 0.52 & 0.46 \\ \midrule
CAOS$_{T/X}$--MiniLM-L6   & \textbf{0.95} 
                               & 0.93 & 0.86 \\
CAOS$_{X/K}$--MiniLM-L6 & 0.84 & 0.83 & \textbf{0.93} \\
CAOS$_{avg}$--MiniLM-L6   & 0.43 & \textbf{0.45} 
                                      & 0.42 \\ \bottomrule  
\end{tabular}
\end{threeparttable}
\caption{Average results for LLaVA, mPLUG-Owl, and MiniGPT-4 across 14 different instructions.}
\label{tab:14_instructions}
\end{table}

\section{Conclusion and Limitations}

Existing methods do not investigate the interplay between object statistics and the semantics of objects in query images or generated context, leaving a gap in understanding object hallucinations during the generation process. To address this, we propose the novel CAOS framework for systematically evaluating object hallucination in LVLMs using generated captions. CAOS focuses on the interaction between generated objects (hallucinated or otherwise), their positional dynamics, ground-truth objects, and object statistics from training data to offer a deeper understanding of hallucination dynamics. Our key contributions include detecting out-of-domain hallucinated objects using LLMs and an oracle based on an ensemble of LVLMs, analyzing sequential generation dynamics, and employing word embedding models to explore the semantic relationships behind hallucinations. We conduct experiments with several diverse LVLMs and find that CAOS effectively identifies hallucinations and provides insights into trade-offs, such as the tendency of certain models to hallucinate frequent objects from training datasets versus ground-truth-related objects. Notably, MiniGPT-4 demonstrates competitive performance across metrics, suggesting that it tends to hallucinate fewer and more contextually relevant objects. In summary, CAOS provides a systematic and nuanced framework for understanding hallucination dynamics, supporting the development of more reliable and robust LVLMs.

It is important to note the limitations of our study despite the extensive exploration undertaken. We focus only on object hallucination in LVLMs, leaving out other performance aspects such as the ability to generate more articulate or contextually coherent responses. Moreover, we use a partial validation set of 2000 MSCOCO images due to computational constraints, which could potentially skew our results. However, for consistency with existing works, we retained the same subset of images employed by \citeauthor{li2023evaluating} (\citeyear{li2023evaluating}). Additionally, our reliance on rule-based object detection, augmented by an LLM and an oracle for verification, may occasionally lead to inaccuracies due to errors in any of these components, though such cases are likely rare. Finally, our analysis considers only a small subset of state-of-the-art LVLMs, excluding some newer or closed-source models. Nevertheless, we view these findings as a step forward in developing more reliable and human-aligned LVLMs. Future work could extend the CAOS framework to encompass other types of hallucinations, such as spatial, relational, or numerical inconsistencies, offering a holistic evaluation of an LVLM's multimodal understanding.



\bibliography{aaai25}

\newpage
\appendix
\onecolumn

\section{List of Instructions}

The list of all instructions used for our experiments is detailed in Table \ref{tab:captions}.

\begin{table*}[!h]
\centering
\begin{tabular}{l|l}
\toprule
Sl. No. & \multicolumn{1}{c}{Instruction}                                              \\
\midrule
1.      & "Provide a brief description of the given image."                            \\
2.      & "Question: Generate a short caption of the image. Answer: "                  \\
3.      & "Create a short textual summary for the image."                              \\
4.      & "Generate a concise description for the image."                              \\
5.      & "Write a succinct summary capturing the essence of the image."               \\
6.      & "Craft a brief narrative that encapsulates the scene depicted in the image." \\
7.      & "Summarize the image with a few descriptive words."                          \\
8.      & "Compose a short, evocative caption for the image."                          \\
9.      & "Describe the image using minimal words but maximum impact."                 \\
10.     & "Formulate a concise and descriptive caption for the image."                 \\
11.     & "Write a short, impactful description for the image."                        \\
12.     & "Sum up the image in a few words, capturing its essence effectively."        \\
13.     & "Craft a brief but descriptive caption for the image."                       \\
14.     & "Write a concise summary that encapsulates the image's message or mood."     \\
\bottomrule
\end{tabular}
\caption{List of all instructions used for our experiments.}
\label{tab:captions}
\end{table*}

\section{Details of models used}

All the evaluated LVLMs used in our experiments consist of three main parts: a vision encoder (VE), a large language model (LLM), and lastly an alignment model (AN) meant to serve as a common mapping between the VE and the LLM. We report a comparison between the LVLMs in Table \ref{tab:comparison}, comparing the size of the models, and the respective pretraining and finetuning recipes (on visual instruction data).

\begin{table*}[!h] 
\centering
\small
\begin{tabular}{llllcccccccc}
\toprule
\multirow{2.5}{*}{\textbf{Model}} &\multirow{2.5}{*}{VE} & \multirow{2.5}{*}{AN} & \multirow{2.5}{*}{LLM} &\multicolumn{3}{c}{Pre-training} & \multicolumn{3}{c}{Fine-tuning}\\
\cmidrule{5-7} \cmidrule{8-10}
  & & & & VE & AN & LLM  & VE & AN & LLM  \\
\midrule
InstructBLIP    & ViT-G/14 & Q-Former  & $\mbox{Vicuna}_{\mbox{\small 13B}}$ & Frozen  & Trained & Frozen & Frozen & Trained & Frozen  \\ 
LLaVA           & ViT-L/14 & Linear    & $\mbox{LLaMA}_{\mbox{\small 13B}}$  & Frozen  & Trained & Frozen & Frozen & Trained & Trained \\
mPLUG-Owl       & ViT-L/14 & Attention & $\mbox{LLaMA}_{\mbox{\small 7B}}$   & Trained & Trained & Frozen & Frozen & Frozen  & LoRA    \\
MiniGPT-4       & ViT-G/14 & Linear    & $\mbox{Vicuna}_{\mbox{\small 13B}}$ & Frozen  & Trained & Frozen & Frozen & Trained & Frozen  \\
Multimodal-GPT  & ViT-L/14 & Attention & $\mbox{LLaMA}_{\mbox{\small 7B}}$   & Frozen  & Trained & Frozen & Frozen & Frozen  & LoRA    \\
\bottomrule
\end{tabular}
\caption{Comparison of the evaluated LVLMs: ``Trained" denotes full pretraining or finetuning, while ``LoRA" denotes that low-rank adapters were trained with the backbone frozen.} 
\label{tab:comparison}
\end{table*}

\section{Out-of-domain object verification using the Oracle}

We further illustrate the performance of the oracle on captions generated by MultimodalGPT (which is not part of the ensemble of LVLMs used as the oracle) for a few example images in Table \ref{tab:images}. The illustration consists of successful, ambiguous (such as ``sun" in the 3rd example which can be interpreted as both the star, which is absent, or sunlight, which is present) as well as erroneous decisions (``apron" in the 5th example image) made by the oracle.

\begin{table*}
\centering
\begin{tabular}{c|p{3.7cm}|p{3.7cm}}
\toprule
Image & Detected out-of-domain objects (with actual tags for presence/absence) & Oracle decisions \\
\midrule
$\includegraphics[scale=0.17]{examples/img1.jpg}$ & \makecell[bl]{lettuce: \textcolor{orange}{\textbf{absent}} \\
                                                                  ingredients: \textcolor{violet}{\textbf{present}} \\
                                                                  tomato: \textcolor{violet}{\textbf{present}} \\
                                                                  onion: \textcolor{orange}{\textbf{absent}}}                &  
                                                    \makecell[bl]{lettuce: \textcolor{orange}{\textbf{absent}} \\
                                                                  ingredients: \textcolor{violet}{\textbf{present}} \\
                                                                  tomato: \textcolor{violet}{\textbf{present}} \\
                                                                  onion: \textcolor{orange}{\textbf{absent}}}                \\ \midrule
$\includegraphics[scale=0.17]{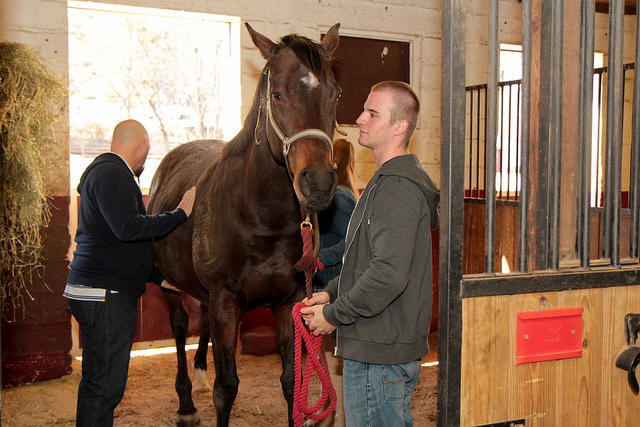}$ & \makecell[bl]{stall: \textcolor{violet}{\textbf{present}} \\
                                                                  clipboard: \textcolor{orange}{\textbf{absent}} \\
                                                                  shirt: \textcolor{violet}{\textbf{present}} \\
                                                                  pants: \textcolor{violet}{\textbf{present}} \\
                                                                  barn: \textcolor{violet}{\textbf{present}}}                &  
                                                    \makecell[bl]{stall: \textcolor{violet}{\textbf{present}} \\
                                                                  clipboard: \textcolor{orange}{\textbf{absent}} \\
                                                                  shirt: \textcolor{violet}{\textbf{present}} \\
                                                                  pants: \textcolor{violet}{\textbf{present}} \\
                                                                  barn: \textcolor{violet}{\textbf{present}}}                \\ \midrule
$\includegraphics[scale=0.17]{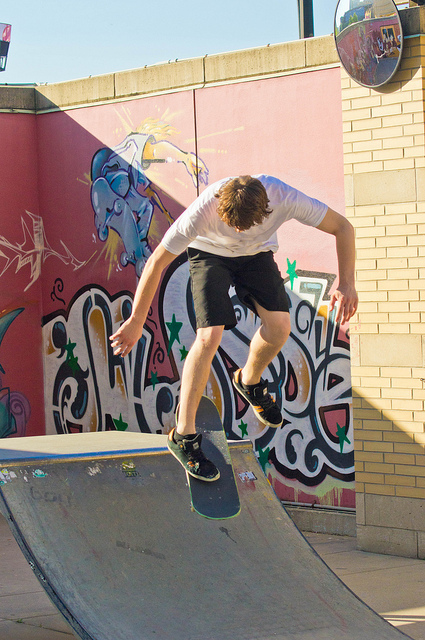}$ & \makecell[bl]{jeans: \textcolor{orange}{\textbf{absent}} \\
                                                                  wall: \textcolor{violet}{\textbf{present}} \\
                                                                  t-shirt: \textcolor{violet}{\textbf{present}}}             &  
                                                    \makecell[bl]{jeans: \textcolor{orange}{\textbf{absent}} \\
                                                                  wall: \textcolor{violet}{\textbf{present}} \\
                                                                  t-shirt: \textcolor{violet}{\textbf{present}}}             \\ \midrule
$\includegraphics[scale=0.17]{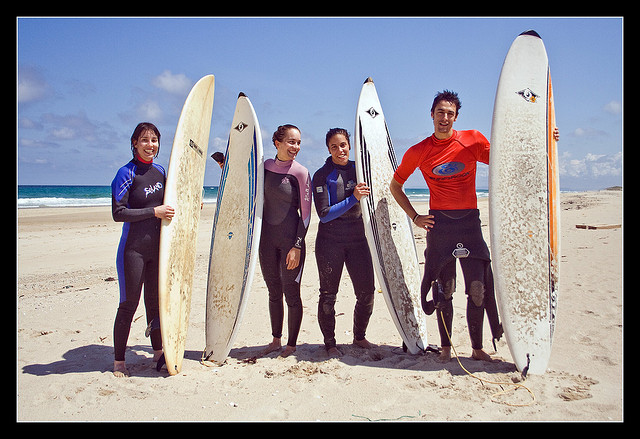}$ & \makecell[bl]{sun: \textbf{ambiguous} \\
                                                                  beach: \textcolor{violet}{\textbf{present}} \\
                                                                  rock: \textcolor{orange}{\textbf{absent}}}                 &  
                                                    \makecell[bl]{sun: \textcolor{violet}{\textbf{present}} \\
                                                                  beach: \textcolor{violet}{\textbf{present}} \\
                                                                  rock: \textcolor{orange}{\textbf{absent}}}                 \\ \midrule
$\includegraphics[scale=0.17]{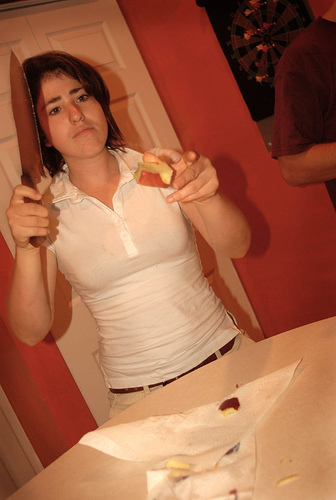}$ & \makecell[bl]{apron: \textcolor{orange}{\textbf{absent}} \\
                                                                  pineapple: \textcolor{orange}{\textbf{absent}}}            &  
                                                    \makecell[bl]{apron: \textcolor{violet}{\textbf{present}} \\
                                                                  pineapple: \textcolor{orange}{\textbf{absent}}}            \\
\bottomrule
\end{tabular}
\caption{Examples of out-of-domain object verification by the oracle.}
\label{tab:images}
\end{table*}

\end{document}